\documentclass[10pt,twocolumn,letterpaper]{article}

\usepackage{cvpr}              % To produce the CAMERA-READY version

\usepackage[utf8]{inputenc}
\usepackage{makecell}
\usepackage{graphicx}
\usepackage{microtype}
\usepackage{braket}
\usepackage{amsmath}
\usepackage{mathtools}
\usepackage{bm}
\usepackage{booktabs}
\usepackage{multirow}
\usepackage[table,xcdraw]{xcolor} % for \cellcolor
\usepackage{graphicx}      % for \resizebox
\usepackage[accsupp]{axessibility}
\definecolor{cvprblue}{rgb}{0.21,0.49,0.74}
\usepackage[pagebackref,breaklinks,colorlinks,allcolors=cvprblue]{hyperref}

\def\paperID{*****} % *** Enter the Paper ID here
\def\confName{CVPR}
\def\confYear{2026}

\title{Image Classification Using CNN-QNN Hybrid Model with Optimized Correlated Features}

\author{
Minseo Seong \quad Youngwook Kim\\
Sogang University\\
{\tt\small r0sa@sogang.ac.kr \quad youngkim@sogang.ac.kr}
}

\begin{document}

\maketitle

\begin{abstract}
We propose a method to optimize the correlation among convolutional neural network (CNN) features that are used as inputs to quantum neural network (QNN) to enhance image classification accuracy. Unlike prior approaches that employ orthogonal decomposition as preprocessing, we intentionally introduce correlated features that are more physically compatible with QNN. This design leverages the QNN’s inherent ability to exploit quantum entanglement for representing correlated states---an advantage unavailable to classical neural networks. We hypothesize that aligning feature correlations with the entanglement structure of QNN improves binary classification performance. Based on a mathematical derivation of QNN outputs, Monte Carlo simulations indicate that an average correlation between features of 0.5 yields optimal classification accuracy. To validate this finding, we evaluate a quantum--classical hybrid model on three tasks: CIFAR-10 (automobile vs. truck), Fashion-MNIST (shirt vs. coat), and radar micro-Doppler signatures (robotic dogs vs. non-robots). To regulate feature correlations, we introduce a correlation-regularization term on the outputs of the CNN, driving the off-diagonal entries of the feature correlation matrix toward a target constant. Across all datasets, inducing intermediate correlation consistently improved accuracy compared to low, high, or unregulated correlations, while also reducing classification accuracy variance. These results demonstrate that imposing moderate feature correlations---without modifying the quantum circuit—enhances classification accuracy and stability by aligning feature statistics with the QNN’s entanglement structure. This study highlights the potential of QNN to surpass the performance of classical classifiers as more qubits become available.  
\end{abstract}    
\section{Introduction}
\label{sec:intro}

Recent progress in quantum neural network (QNN), which utilizes quantum phenomena such as superposition and entanglement, has led to hybrid models combining classical feature extractors like convolutional neural network (CNN) and QNN classifiers for image recognition. This architecture leverages the expressive power of quantum representations while operating under the constraints of noisy intermediate scale quantum (NISQ) devices with limited number of qubits \cite{preskill2018quantum}. This model has the potential to outperform classical deep learning methods for image classification \cite{hafeez2024h, chalumuri2022quantum}.

When implementing QNN using quantum circuits, deep circuits with many quantum gates---analogous to layers in deep learning---are discouraged. This is because increasing circuit depth accumulates noise and reduces coherence \cite{cheng2021simulating}, and may lead to barren plateaus, a vanishing gradient in QNN \cite{mcclean2018barren}. To enhance QNN performance with reduced circuit depth, utilizing quantum entanglement enhances the ability to represent quantum states \cite{sim2019expressibility}. Therefore, it is desirable to improve shallow, qubit-limited QNN without increasing circuit depth by better utilizing quantum entanglement.

\begin{figure}[t]
  \centering
  \includegraphics[width=1\linewidth]{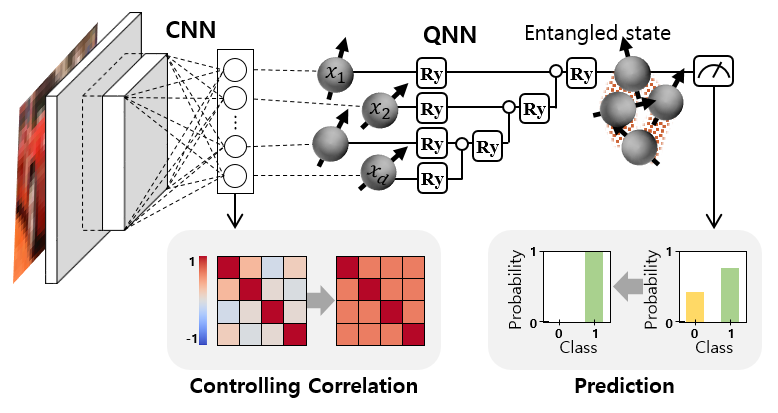}
  \caption{Overview of the proposed model. An input image is processed by a classical CNN to extract 8-dimensional latent features. A correlation regularization module enforces a predefined level of inter-feature correlation before quantum encoding. The features are mapped onto a quantum feature map and a variational quantum circuit. This framework enables improved accuracy and stability by controlling classical feature statistics with the quantum entanglement structure.}
  \label{fig}
\end{figure}

Entanglement produces nonclassical dependencies among qubits, which is the joint state that cannot be written as a product of single-qubit states. In quantum data encodings, each feature is encoded onto a qubit and entangling gates exposes multi-qubit interactions, distinguishing QNN from classical models. We therefore expect that classical and quantum models handle data in different ways. In classical models, orthogonal feature extraction is often applied to reduce redundancy. However, the effect of correlations among classical input features on quantum models, where entanglement is the main resource, remains underexplored.

This study proposes optimizing the feature correlations to improve image classification accuracy in a hybrid CNN--QNN model. Conventionally, features obtained from the orthogonal decomposition of input data have been used as inputs to QNN \cite{jerbi2023quantum, hur2022quantum, gong2024quantum}, ensuring zero correlation between features. In contrast, we hypothesize and validate that moderate correlation among classical input features is necessary to harness the idiosyncrasies of using QNN. Intuitively, entanglement induces interactions between qubits to induce correlation between them. Because the states of the qubits are determined by the value of the classical feature, quantum states are affected by classical correlations among features. If inputs are aggressively decorrelated, entangled states may carry little information, whereas excessive correlation collapses effective dimensionality and increases redundancy.  Thus, introducing a moderate correlation provides a balance between them. This mirrors the transition effect reported by Wang \etal \cite{wang2024transition}: with limited measurement shots, increasing entanglement---quantified by Schmidt rank---first reduces error and then increases it, yielding an intermediate optimum. 

We suggest a correlation-regularized hybrid CNN--QNN architecture that explicitly optimizes input feature correlations to maximize classification accuracy. Our model consists of a CNN feature extractor followed by a QNN classifier implemented as a phase-encoded variational quantum circuit (VQC). We propose a covariance-based correlation loss that forces off-diagonal elements of the feature covariance matrix toward a target value, thereby encouraging controlled feature correlation. To evaluate the hybrid model’s performance, we first mathematically analyze the relationship between feature correlation and classification accuracy on random data. Then  binary classification experiments are conducted on the Fashion-MNIST and CIFAR-10 image datasets, as well as on radar micro-Doppler signatures. As a result, we verify that an intermediate level of feature correlation enhances the performance of CNN–QNN.

Our contributions are summarized as follows: (1) a regularization method that aligns input feature correlation with quantum entanglement, (2)  a mathematical analysis and quantification of the influence of input feature correlations on QNN classification performance, revealing an optimal correlation regime, and (3) an empirical demonstration of improved accuracy and stability across three image datasets through inducing intermediate feature correlation.

The remainder of this paper is organized as follows. \cref{sec:related,sec:prelim} respectively review related works and the fundamental concepts of quantum computing and QNN. \cref{sec:method} presents our hybrid CNN--QNN model and details the correlation control mechanism. \cref{sec:anal} provides a numerical analysis of the relationship between input correlation and QNN performance. \cref{sec:veri} describes the experimental setup and results across multiple datasets. \cref{sec:con} concludes the paper with a summary and future directions.

\section{Related Work}
\label{sec:related}

\paragraph{Decorrelation} In classical deep learning, a common strategy for feature extraction is to decorrelate features or weights. This strategy helps reduce redundancy and avoid overfitting, thereby improving generalization in both supervised and self-supervised settings. Falez \etal \cite{falez2020improving} applied whitening methods such as principal component analysis and zero phase component analysis to decorrelate inputs for spiking neural network, and reported improved classification accuracy for the CIFAR-10 dataset over unwhitened baselines. Pan \etal \cite{pan2019switchable} showed that replacing batch/instance normalization with layers applying decorrelating activations yields consistent gains across classification, segmentation, and domain adaptation. In the self-supervised learning area, Barlow Twins \cite{zbontar2021barlow} aligns two augmented views while driving the cross-correlation matrix toward the identity, suppressing off-diagonal terms and thus decorrelating features. Cogswell \etal \cite{cogswell2015reducing} showed that strong correlations among hidden layers in deep networks are closely tied to overfitting, and proposed a method that explicitly minimizes cross-covariance between features to encourage diversity. Similarly, Rodriguez \etal \cite{rodriguez2016regularizing} demonstrated that enforcing local orthogonality among CNN weights effectively reduces overfitting and enhances the model’s generalization performance. However, these decorrelation strategies are largely confined to classical deep learning pipelines. On the quantum side, the effect of classical input correlations on QNN performance has received limited attention. 

\paragraph{Hybrid QNN} Hybrid quantum--classical models have been applied to image classification \cite{hafeez2024h, chalumuri2022quantum, fan2023hybrid, ray2024hybrid, afane2025atp}, stock market forecasting \cite{choudhary2025hqnn}, and radar signal classification \cite{ghosh2024hybrid, liu2025radar}. Across these studies, the emphasis is on architectural combinations (CNN \cite{hafeez2024h, chalumuri2022quantum, ghosh2024hybrid} and graphical neural network \cite{ray2024hybrid} feature extractors with VQC), encoding methods (A Flexible Representation of Quantum Images \cite{fan2023hybrid, le2011flexible} and Adaptive Threshold Pruning \cite{afane2025atp}), and training regimes (frozen vs. end-to-end \cite{ray2024hybrid}). Prior work even avoids entanglement altogether \cite{hafeez2024h}. However the correlation structure of the classical feature vectors fed into a VQC has not been explicitly analyzed or controlled. This gap motivates our correlation-regularized hybrid CNN--QNN: we align classical feature covariance with quantum entanglement to improve shallow, qubit-limited VQCs without increasing circuit depth.

\section{Preliminaries}
\label{sec:prelim}

QNN is composed of quantum gates. A quantum gate that acts on qubits is a basic building block of a quantum circuit. This section outlines the fundamental concepts used in the remainder of this study \cite{nielsen2010quantum}.
%-------------------------------------------------------------------------
\subsection{Qubits}

Quantum computers use quantum bits, or qubits, as their fundamental units of computation. In quantum computing, a qubit can be in a quantum state, which can be a superposition of basis states $\ket{0}$ and $\ket{1}$. Here, quantum state $\ket{\psi}$ is denoted as: $\ket{\psi}=\alpha\ket{0}+\beta\ket{1}$, where $\alpha$ and $\beta$ are complex numbers whose absolute squares represent the likelihood of the qubit being measured in each state. These single-qubit states can also be described in a 2-dimensional column vector:  $\ket{0}= \begin{bsmallmatrix} 1 \\ 0 \end{bsmallmatrix},\ 
 \ket{1} = \begin{bsmallmatrix} 0 \\ 1 \end{bsmallmatrix},\ 
 \ket{\psi} = \begin{bsmallmatrix} \alpha \\ \beta \end{bsmallmatrix}$. By using the Bloch sphere, a single qubit can be expressed geometrically, then $\ket{\psi}$ can be written as: $\ket{\psi} = \cos\!\left(\frac{\theta}{2}\right)\ket{0} 
+ e^{j\phi}\sin\!\left(\frac{\theta}{2}\right)\ket{1}$, where $\theta$ is the polar angle from the $\ket{0}$ state, and $\phi$ is the azimuthal angle.

In a multiple-qubit system, the quantum state of each qubit can exhibit correlations with the others, a phenomenon referred to as quantum entanglement. For example, one of the Bell states that represents maximum entanglement between two qubits is $\frac{1}{\sqrt{2}}(\ket{00} + \ket{11})$. This state means perfectly correlated qubits. If the first qubit is 0, the second is also 0; if the first is 1, the second is 1. In other words, entangled states cannot be written as a product of single-qubit states; they are non-separable and exhibit correlations.

%-------------------------------------------------------------------------
\subsection{Quantum Gate}
Similar to logic gates in classical computers, quantum computers operate using quantum gates. Key examples include the Hadamard gate, Pauli gates, rotation gates, and controlled not (CNOT) gate. (i) Hadamard gate initializes a qubit into a superposition. (ii) Pauli gates include four types of gates: identity, Pauli-X, Pauli-Y, and Pauli-Z gate, defined in \cref{eq:gates}.

\begin{equation}
\begin{aligned}
I &= \begin{bmatrix} 1 & 0 \\ 0 & 1 \end{bmatrix}, &
X &= \begin{bmatrix} 0 & 1 \\ 1 & 0 \end{bmatrix}, \\[0.3em] 
Y &= \begin{bmatrix} 0 & -j \\ j & 0 \end{bmatrix}, &
Z &= \begin{bmatrix} 1 & 0 \\ 0 & -1 \end{bmatrix}
\end{aligned}
\label{eq:gates}
\end{equation}
The identity gate does not change the state of a qubit; The Pauli-X gate flips the state of a qubit between $\ket{0}$ and $\ket{1}$; The Pauli-Y gate also flips the qubit state, but introduces a phase difference, combining both bit-flip and phase-flip operations; The Pauli-Z gate applies a phase flip to the $\ket{1}$ component, leaving $\ket{0}$ unchanged. (iii) Rotation gates implement unitary operations that rotate the state of a single qubit by a defined angle($\theta$) around the $x$-, $y$-, or $z$-axis on the Bloch sphere. They are defined as $R_x(\theta)=e^{-j\frac{\theta}{2}X}$, $R_y(\theta)=e^{-j\frac{\theta}{2}Y}$ and $R_z(\theta)=e^{-j\frac{\theta}{2}Z}$ respectively. (iv) CNOT gate operates on two qubits, flipping the state of the target qubit when the control qubit is in the $\ket{1}$ state. 
%-------------------------------------------------------------------------
\subsection{Quantum Neural Network (QNN)}
QNN aims to find a function $f_\theta(x)$ for an input data sample $x$ that predicts the class of $x$. It is composed of a feature map and a VQC, denoted by $U(x)$ and $W(\theta)$, respectively. Both the feature map and VQC are sequences of quantum gates. The feature map embeds classical data $x$ into angles of rotation gates, thereby encoding it into quantum states. VQC includes rotation gates parameterized by angles $\theta$, which serve as trainable parameters and are optimized iteratively according to a predefined objective function. A value for $f_\theta(x)$ corresponds to the expectation value of an observable $\hat{H}$ with respect to quantum states prepared by quantum circuits, represented as:
\begin{equation}
\ket{\psi(\theta)}=W(\theta)U(x)\ket{0}.
\label{eq:phi_theta}
\end{equation}
The observable $\hat{H}$ represents any measurable physical quantity from a qubit. It is a Hermitian operator expressed as linear combination of Pauli operators. As a result, the function $f_\theta(x)$ can be written as \cref{eq:f_theta1}.

\begin{equation}
\begin{aligned}
f_\theta(x)
&= \braket{\psi(\theta) | \hat{H} | \psi(\theta)} \\[0em]
&= \braket{0 | H\, U^{\dagger}(x)\, W^{\dagger}(\theta)\,
            \hat{H}\, W(\theta)\, U(x)\, H | 0}
\end{aligned}
\label{eq:f_theta1}
\end{equation}

When it comes to optimizing the parameters of the VQC, backpropagation using automatic differentiation cannot be directly applied to QNN. This is due to the nature of quantum measurement, which irreversibly collapses the quantum state and makes reuse impossible. To estimate the gradient of expectation value with respect to a parameter of quantum gate, the parameter-shift rule \cite{schuld2019evaluating} is commonly used. This rule allows the computation of exact gradients for specific gates by evaluating the quantum circuit at shifted parameter values: $\frac{\partial g(\mu)}{\partial \mu} = r \left[ g(\mu + s) - g(\mu - s) \right]$, where $s = \frac{\pi}{4r}$. Here $g(\mu)$ is the expectation value of a Hermitian observable measured on a quantum state prepared with parameter $\mu$, and $r$ is the magnitude of the eigenvalues of the Hermitian generator of the quantum gate. This method enables gradient estimation without requiring access to internal quantum states or backpropagation through the quantum layer.
\section{Proposed Methodology}
\label{sec:method}

\begin{figure}[t]
  \centering
  \includegraphics[width=0.9\linewidth]{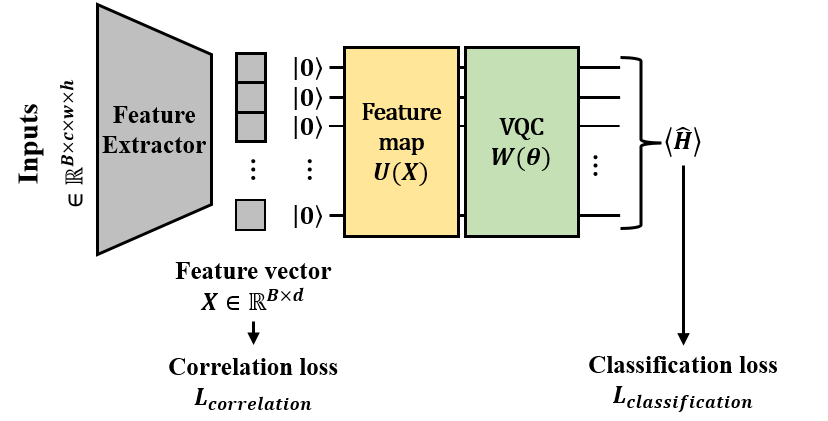}
  \caption{Proposed framework.}
  \label{fig2}
\end{figure}

In our study, a combination of CNN for feature extraction and QNN for classification is employed as shown in \cref{fig2}. Feature extractor extracts features from an input image, then the features are encoded to quantum states using a quantum feature map, followed by a VQC and an estimator to predict the label of the input image. Beyond a standard hybrid CNN–QNN, we add a correlation regularizer to the CNN output. The regularizer steers the batch correlation matrix of the latent features toward a predefined target. 

%-------------------------------------------------------------------------
\subsection{Structure of Suggested Quantum Classifier}

\begin{figure}[t]
  \centering
  \includegraphics[width=0.9\linewidth]{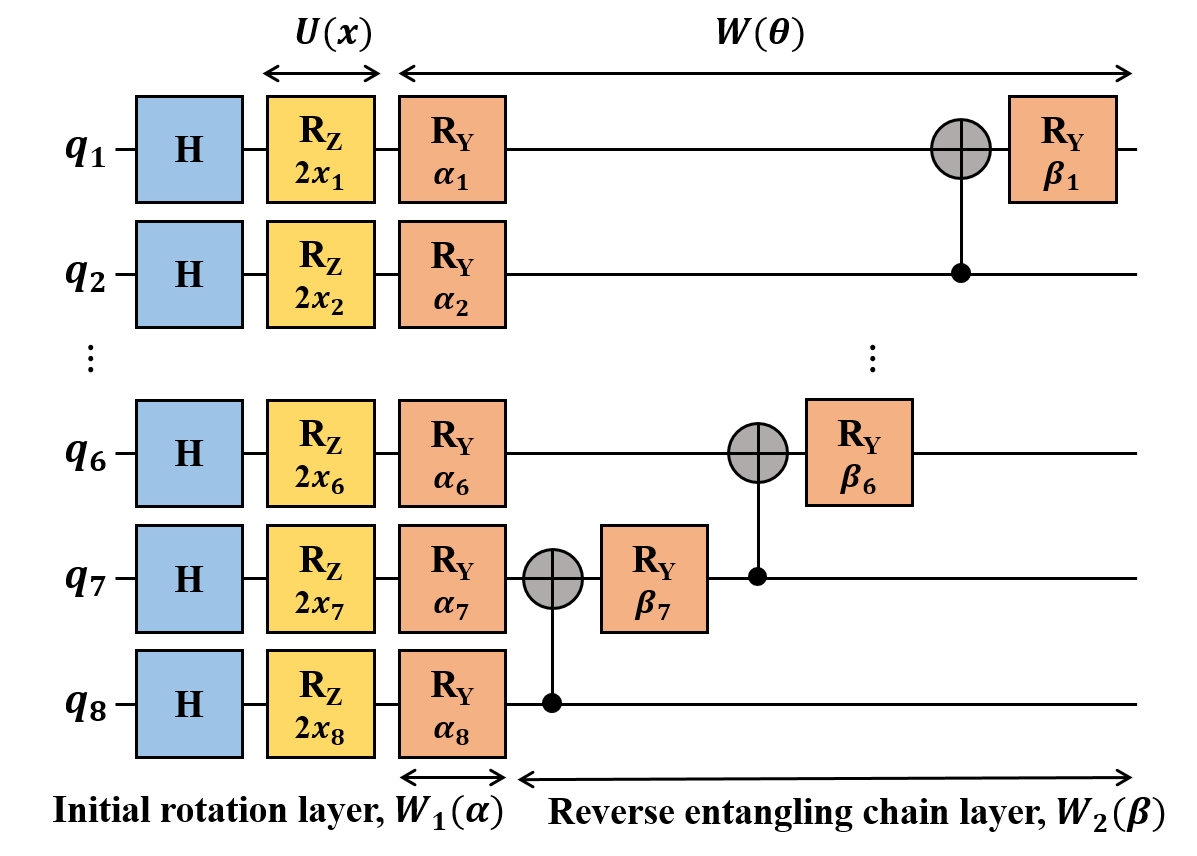}
  \caption{Quantum circuit configuration.}
  \label{fig3}
\end{figure}

We consider a $d$-qubit quantum classifier composed of two stages as shown in \cref{fig3}: a feature map $U(X)$ and a VQC $W(\theta)$. Let $X = [x_j^{(i)}]_{i=1,j=1}^{B,d} \in \mathbb{R}^{B \times d},\; x^{(i)} \in \mathbb{R}^d$ where $B$ is the batch size and $\textit{d}$ is the dimensionality of the classical latent vector extracted from the classical CNN encoder. First, all qubits are initialized to $\ket{0}$ states, followed by applying the Hadamard gate to prepare $\ket{+}=\frac{1}{\sqrt2}(\ket{0}+\ket{1})$ states. For the next step, we adopt the ZFeatureMap \cite{havlivcek2019supervised} as:
\begin{equation}
U(x^{(i)}) = \prod_{j=1}^{d} R_z\left(2x_j\right)^{(j)}
\label{eq:ZFeat}
\end{equation}  where $R_z(\,\cdot\,)^{(q)}$ is the single qubit Z-rotation acting on qubit $q$, $Z_q$ denotes the Pauli-Z gate acting on qubit $q$. Thus, $U(X)$ encodes each classical feature $x_j$ into the phase of the corresponding qubit via a Z-rotation.

This study proposes composing the $W(\theta)$ of VQC into two parts as depicted in \cref{fig3}: initial rotation layer and reverse entangling chain layer. Let $\theta= \allowbreak (\alpha_1,\dots,\alpha_d,\allowbreak \beta_1,\dots,\beta_{d-1})$ be the set of trainable parameters of the VQC. Here $\alpha_j$ corresponds to a single-qubit rotation angle in the initial rotation layer, and $\beta_k$ corresponds to the rotation applied during the reverse entangling chain layer. We define the initial rotation layer as:
\begin{equation}
W_1(\alpha)
= \prod_{j=1}^{d} R_y\left(\alpha_j\right)^{(j)}
\label{eq:ZFeatureMap}
\end{equation} which rotates each qubit about $y$-axis by an angle $\alpha$. The reverse entangling chain layer is defined by \cref{eq:W2}. 
\begin{equation}
W_2(\beta)
= \prod_{k=1}^{d-1} R_y\left(\beta_k\right)^{(k)}\text{CNOT}_{(k+1)\rightarrow(k)}
\label{eq:W2}
\end{equation}
Here we denote the CNOT gate with control qubit $c$ and target qubit $t$ as $\text{CNOT}_{c\rightarrow t}$. This layer applies CNOT gates and $Ry$ rotations from qubits with higher indices to those with lower indices, compressing the information into the first qubit. The variational layer $W(\theta)$ is given by:
\begin{equation}
W(\theta)=W_2(\beta)W_1(\alpha)
\label{eq:W}
\end{equation}

This architecture is designed to progressively accumulate correlations toward the first qubit. The CNOT gate controls the state of target qubit based on the state of control qubit. Therefore, the target qubit captures information of the control qubit. For our VQC, the CNOT gates are arranged in reverse order, from higher-index qubits toward lower-index qubits. Thus multi-qubit correlations are generated and gradually transferred to the first qubit during the reverse entangling chain layer. Under this design, measuring a single-qubit observable $\hat{H} = Z \otimes I^{\otimes (d-1)} \coloneqq Z_1$  on the first qubit suffices to extract information about the entire system state, reducing shot complexity and training time. Consequently, the overall parameterized quantum state is given by:
\begin{equation}
\begin{gathered}
f_\theta(X) = \bra{\psi(\theta)} Z_1 \ket{\psi(\theta)}, \\[0.em]
\text{where } \ket{\psi(\theta)} = W(\theta)\, U(X)\, H\ket{0}
\end{gathered}
\label{eq:vqc-output}
\end{equation}

We adopt the negative log-likelihood loss for binary classification. Because the expectation value of any observable in quantum mechanics takes a value between -1 and 1, we rescale $f_\theta(x^{(i)} )$ into probability $p_i=\frac{1+f_\theta (x^{(i)}) }{2}$. Given the label $y_i \in \{0,1\}$, the classification loss is defined as: 
$L_{\text{classification}}=-\frac{1}{B}\sum_{i=1}^B(y_i\log{(p_i)}+(1-y_i)\log{1-(p_i)}).$

%-------------------------------------------------------------------------
\subsection{Control of Feature Correlation}
\begin{figure}[t]
    \centering

    \begin{minipage}{0.33\linewidth}
        \centering
        \subfloat[$Cor=0.0$]{\includegraphics[width=\linewidth]{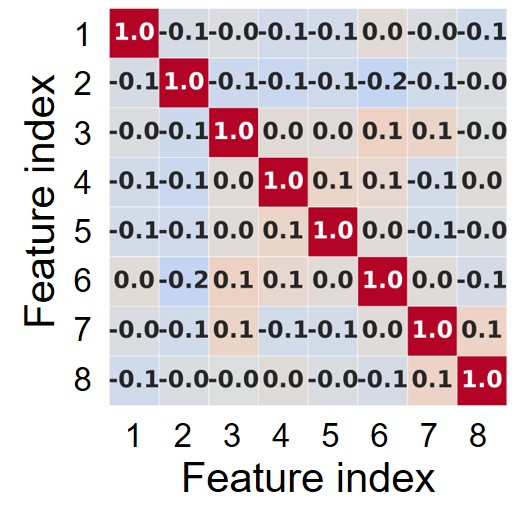}\label{fig:fig7a}}
    \end{minipage}\hfill
    \begin{minipage}{0.33\linewidth}
        \centering
        \subfloat[$Cor=0.5$]{\includegraphics[width=\linewidth]{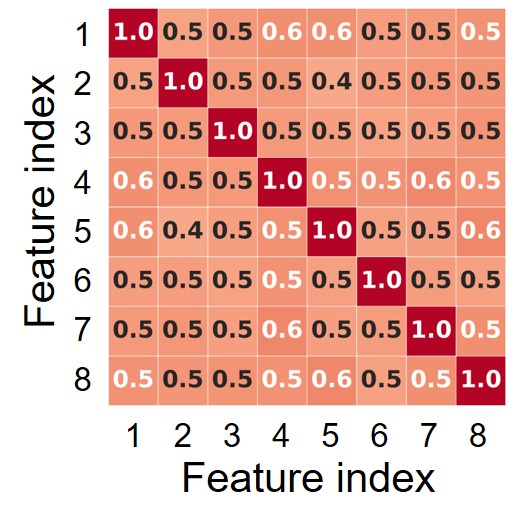}\label{fig:fig7b}}
    \end{minipage}\hfill
    \begin{minipage}{0.33\linewidth}
        \centering
        \subfloat[$Cor=1.0$]{\includegraphics[width=\linewidth]{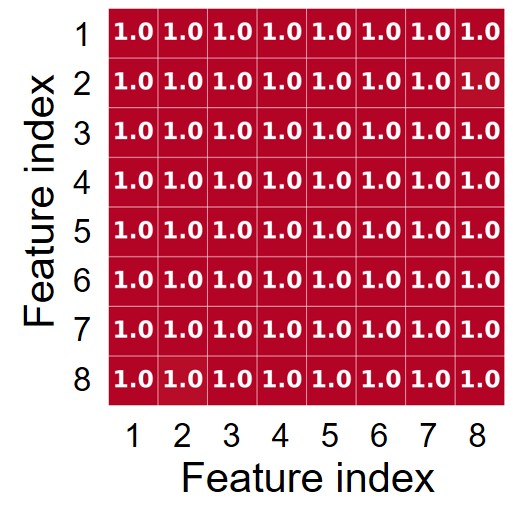}\label{fig:fig7c}}
    \end{minipage}

    \caption{Examples of feature correlation matrices targeting (a) $Cor=0.0$, (b) $Cor=0.5$, and (c) $Cor=1.0$.}
    \label{fig7}
\end{figure}

To investigate the effect of correlation of input feature from CNN on QNN performance, we proposed introducing a correlation loss that enforces a predefined correlation structure among the input latent vectors. To calculate covariance-based correlation matrix, we normalize the features to have zero mean and unit variance across the batch. The empirical correlation matrix $C\in \mathbb{R}^{d \times d}$ is then computed as: $C=\frac{1}{B-1} X_\text{{norm}}^T X_\text{{norm}}$, where $X_\text{{norm}}$ is the normalized $X$. We define a target correlation $T\in \mathbb{R}^{d \times d}$ where all diagonal elements are fixed to 1 and all off-diagonal elements are set to a constant $\mathit{Cor}$ that is a correlation coefficient. The correlation loss is computed as the mean squared error between the empirical and target correlation matrices: 
\begin{equation}
L_{\text{correlation}}=\frac{1}{d^2}  \sum_{i=1}^d\sum_{j=1}^d(C_{ij}-T_{ij} )^2. 
\end{equation}
Examples of correlation matrices computed on the test dataset are shown in \cref{fig7}. Each matrix results from training with a target correlation  $Cor=\{0.0, 0.5, 1.0\}$ and the corresponding target matrix  $T$. During training, the CNN–QNN model is optimized end-to-end using a total loss $L=L_{\text{classification}}+L_{\text{correlation}}$. The empirical correlations do not exactly match the targets due to finite-batch estimation, but the deviations are minor.

\section{Analysis of Correlation-Classification Accuracy Relationship}
\label{sec:anal}

We mathematically analyze how the correlation between input features affects the classification accuracy of the quantum classifier. To examine this effect, we first derive the expression for the QNN output by analyzing the expectation value of a single-qubit observable after passing through the circuit. This is done using a Heisenberg picture formulation, in which the observables vary with time while the states are fixed, unlike the Schrödinger picture \cite{nielsen2010quantum}. We then perform Monte Carlo simulation by calculating the QNN output on random data with different correlation levels. Each dataset is generated with a fixed correlation coefficient $Cor$, and the corresponding classification accuracy is computed. This combined analysis allows us to examine how $Cor$ affects model performance and to identify the correlation range that maximizes accuracy. 

By using the Heisenberg picture, the output of the QNN in \cref{eq:vqc-output} is expressed as:
\begin{equation}
\label{eq:vqc-expand}
\begin{split}
f_\theta(x)
&= \langle + | U(x)^\dagger W_1(\alpha)^\dagger W_2(\beta)^\dagger \\
&\qquad \times Z_1 W_2(\beta) W_1(\alpha) U(x) | + \rangle \\
&= \braket{+ | U(x)^\dagger W_1(\alpha)^\dagger \hat{O}_1 W_1(\alpha) U(x) | +} \\
&= \braket{+ | U(x)^\dagger \hat{O}_2 U(x) | +} \\
&= \braket{+ | \hat{O}_3 | +}
\end{split}
\end{equation}

\noindent where $\hat{O}_1 = W_2(\beta)^\dagger Z_1 W_2(\beta)$, $\hat{O}_2 = W_1(\alpha)^\dagger \hat{O}_1 W_1(\alpha)$, and $\hat{O}_3 =U(x)^\dagger \hat{O}_2 U(x)$. The evaluation reduces to computing the expectation of the effective observable $\hat{O}_3$ with respect to $\ket{+}$. Prior to the main derivation, there are a few identities to use \cite{nielsen2010quantum}:
\begin{alignat}{2}
    & R_y(\theta_j)^{(i)\dagger} Z_i R_y(\theta_j)^{(i)} &&= Z_i \cos(\theta_j) + X_i \sin(\theta_j) \notag \\
    & &&\coloneqq Z_i c_j - X_i s_j \label{eq:ryz} \\
    & R_y(\theta_j)^{(i)\dagger} X_i R_y(\theta_j)^{(i)} &&\coloneqq X_i c_j + Z_i s_j \label{eq:ryx} \\
    & R_z(\theta_j)^{(i)\dagger} X_i R_z(\theta_j)^{(i)} &&\coloneqq X_i c_j - Y_i s_j \label{eq:rzx} \\
    & \text{CNOT}_{(c\rightarrow t)}^{\dagger} Z_t\, \text{CNOT}_{(c\rightarrow t)} &&= Z_c Z_t \label{eq:cnotz} \\
    & \text{CNOT}_{(c\rightarrow t)}^{\dagger} X_t\, \text{CNOT}_{(c\rightarrow t)} &&= X_t \label{eq:cnotx}
\end{alignat}

Using these identities, steps to evaluate effect of $Cor$ on classification accuracy are as follows: step 1 derives $\hat{O}_1$, step 2 derives $\hat{O}_2$, step 3 derives $\hat{O}_3$, step 4 derives the closed-form expression for $f_\theta(x)$; and step 5 describes the Monte Carlo simulation used to evaluate classification accuracy based on $f_\theta(x)$.

\textbf{Step 1: }\bm{$\hat{O}_1$}  To derive $\hat{O}_1$, we examine \cref{eq:vqc-expand} by increasing $k$ and capturing the general pattern. For $k=1$, which means applying $\text{CNOT}_{(2)\rightarrow (1)}$ followed by $R_y (\beta_1 )^{(1)}$, $\hat{O}_1$ for $k=1$ is expressed as: 
\begin{equation}
\begin{aligned}
\hat{O}_1^{(k=1)}
&=\text{CNOT}^{\dagger}_{(2)\rightarrow(1)}R_y(\beta_1)^{(1)\dagger}Z_1\cdots \\[0.em] &\quad R_y(\beta_1)^{(1)}\text{CNOT}_{(2)\rightarrow(1)} 
\label{eq:o1k1_v1}
\end{aligned}
\end{equation}
After that it simplifies to:
\begin{equation}
\begin{aligned}
\hat{O}_1^{(1)}&=\text{CNOT}^{\dagger}_{(2)\rightarrow(1)}(Z_1c_1-X_1s_1)\text{CNOT}_{(2)\rightarrow(1)} \\[0.em]
&=c_1Z_2Z_1-s_1X_1
\label{eq:o1k1_v2}
\end{aligned}
\end{equation} according to \cref{eq:ryz,eq:cnotz,eq:cnotx}. For $k=2$, a similar procedure is applied, with the difference that it only influences qubits 2 and 3. Consequently, we obtain: $\hat{O}_1^{(2)}=c_1c_2Z_3Z_2Z_1-c_1s_2X_2Z_1-s_1X_1$, and with the same procedure, $\hat{O}_1^{(3)}=c_1c_2c_3Z_4Z_3Z_2Z_1-c_1c_2s_3X_3Z_2Z_1-c_1s_2X_2Z_1-s_1X_1$. Therefore, the general form can be captured as:
\begin{equation}
\begin{aligned}
\hat{O}_1= &\left(\prod_{j=1}^{d-1}c_j\right)Z_dZ_{d-1}\cdots Z_1 \\[0em]
&-\sum_{r=1}^{d-1}\left[\left(\prod_{j=1}^{r-1}c_j\right)s_rX_rZ_{r-1}\cdots Z_1\right]
\label{eq:o1}
\end{aligned}
\end{equation}

\textbf{Step 2: }\bm{$\hat{O}_2$} For $\hat{O}_2=W_1(\alpha)^{\dagger}\hat{O}_1W_1(\alpha)$, we analyze the first and second terms of \cref{eq:o1} separately. For the first term of $\hat{O}_1$ in \cref{eq:o1}, $W_1(\alpha)$ applies $R_y(\alpha_1)^{(1)}, R_y(\alpha_2)^{(2)}, \cdots$ to each $Z_1, Z_2, \cdots$ independently, to be $(\prod_{j=1}^{d-1}c_j)\prod_{j=1}^{d}(Z_j\cos(\alpha_j)-X_j\sin(\alpha_j))$ according to \cref{eq:ryz}. However, we prune Pauli strings containing at least one Z at the $\hat{O}_2$ stage in advance as:
\begin{equation}
\begin{aligned}
\left( \prod_{j=1}^{d-1}
\cos(\beta_j)\,(-\sin(\alpha_j))\,X_j \right)
(-\sin(\alpha_d))\,X_d
\label{eq:o2v1}
\end{aligned}
\end{equation} The reason is that in the next step we apply only Z-rotations  $U(X)=\prod_{j=1}^{d}R_z(2x_j)^{(j)}$, which do not change any $Z$ (i.e. $R_z(\theta)^\dagger ZR_z(\theta)=Z$). Therefore any $Z_j$ remains $Z_j$ afterward. Since the final state is $\ket{+}$ and $\bra{+}Z\ket{+}=0$, every Pauli string containing a $Z_j$  contributes zero and can be discarded. In the second term of $\hat{O}_1$, $r_{th}$ summation term is transformed into:
\begin{equation}
\begin{aligned}
\left(\prod_{j=1}^{r-1} c_j\right) s_r
&\left( X_r \cos(\alpha_r) + Z_r \sin(\alpha_r) \right) 
\times \\[0.em]
&\left( Z_{r-1} \cos(\alpha_{r-1}) - X_{r-1} \sin(\alpha_{r-1}) \right)
\cdots \\[0em]
&\left( Z_{1} \cos(\alpha_{1}) - X_{1} \sin(\alpha_{1}) \right)
\label{eq:o2v2}
\end{aligned}
\end{equation}
Again, we can prune these as follows.
\begin{equation}
\label{eq:o2v3}
\begin{split}
\sum_{r=1}^{d-1} \Biggl[ &\left( \prod_{j=1}^{r-1} \cos(\beta_j)(-\sin(\alpha_j))X_j \right) \times \\
&\qquad \sin(\beta_r)\cos(\alpha_r)X_r \Biggr]
\end{split}
\end{equation}
To sum up, $\hat{O}_2$ is rewritten as \cref{eq:o2}.
\begin{equation}
\begin{aligned}
&\hat{O}_2=A_1X_1\cdots X_d+\sum_{r=1}^{d-1}A_2X_1\cdots X_r \\[0.em]
&\text{where } A_1=\left(\prod_{j=1}^{d-1}\cos(\beta_j)(-\sin(\alpha_j))\right)(-\sin(\alpha_d)), \\[0.em]
&A_2=-\left(\prod_{j=1}^{r-1}\cos(\beta_j)(-\sin(\alpha_j))\right)\sin(\beta_r)\cos(\alpha_r).
\label{eq:o2}
\end{aligned}
\end{equation}

\textbf{Step 3: }\bm{$\hat{O}_3$} By \cref{eq:rzx} and for Pauli-Y is also $\bra{+}Y\ket{+}=0$, $\hat{O}_3 = U(x)^\dagger \hat{O}_2 U(x)$ is rewritten as below:
\begin{equation}
\begin{aligned}
\hat{O}_3&=A_1\prod_{j=1}^d(\cos(2x_j)X_j) \\[0.em]
&+\sum_{r=1}^{d-1}A_2\prod_{j=1}^r(\cos(2x_j)X_j)
\label{eq:o3}
\end{aligned}
\end{equation}

\textbf{Step 4: }\bm{$f_\theta(x)$} Since $\bra{+}X\ket{+}=1$ and $\bra{+}I\ket{+}=I$, 
$f_\theta(x)$ is summarized as:
\begin{equation}
\begin{aligned}
f_\theta(x)=A_1\prod_{j=1}^d\cos(2x_j)+\sum_{r=1}^{d-1}A_2\prod_{j=1}^r\cos(2x_j)
\label{eq:f}
\end{aligned}
\end{equation}
With the derived form of $f_\theta(x)$, the output of the model can be computed in closed form for any given parameters $\alpha, \beta$ and inputs $x$.

\textbf{Step 5: Investigation of Classification Accuracy Depending on Classical Feature Correlation} Using the $f_\theta(x)$ in \cref{eq:f}, we simulate the classification accuracy on random data samples with various correlation matrices. Under the assumption that the outputs of the feature extractors are approximately normally distributed \cite{garriga2018deep}, synthetic data were generated with various dimensions and two classes. Specifically, samples from Class 1 were drawn from $N(0,T)$, while samples from Class 2 were drawn from $N(\mu,T)$, where $\mu$ varied within ${\{0.6,0.8,\dots,1.40\}}$. Both classes contained an equal number of samples to ensure balanced datasets. Each experiment is repeated five times with different random seeds, and results are averaged. Dataset has 10,000 samples in total for each setting, and they are split into 8:2 for training and test. $\alpha$ and $\beta$ are assumed to be distributed as $N(0,1)$. We treat $f_\theta(x)$ as a one-dimensional score and find a scalar threshold by exhaustive search on sorted training scores. 

\begin{figure}[t]
  \centering
  % 1st row
  \subfloat[$d=8$]{%
    \includegraphics[width=0.49\linewidth]{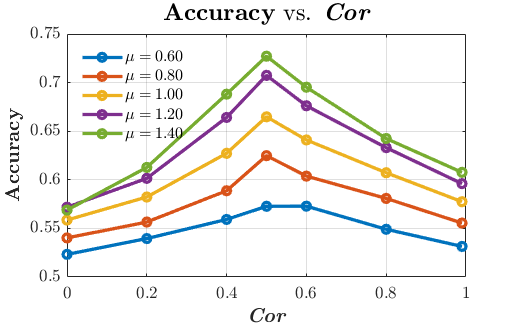}%
  }\hfill
  \subfloat[$d=32$]{%
    \includegraphics[width=0.49\linewidth]{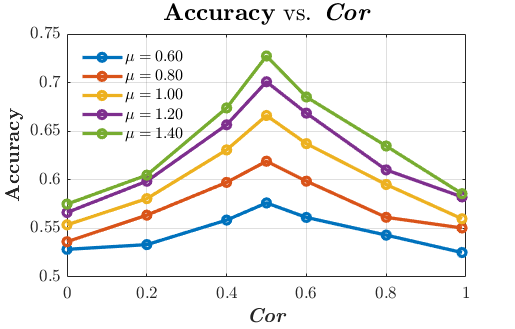}%
  }\\[0em]
  % 2nd row
  \subfloat[$d=100$]{%
    \includegraphics[width=0.49\linewidth]{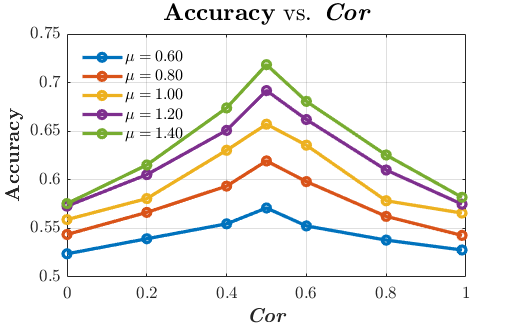}%
  }\hfill
  \subfloat[$d=1000$]{%
    \includegraphics[width=0.49\linewidth]{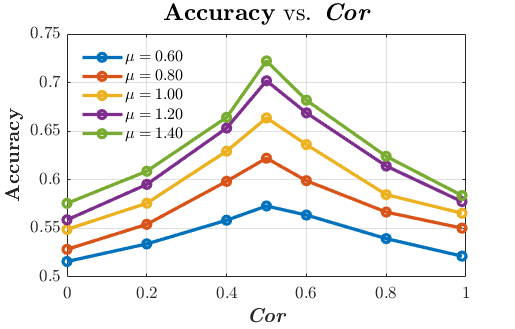}%
  }
  
  \caption{Accuracy vs. $Cor$ across various dimensions $d$.}
  \label{fig4}
\end{figure}

\cref{fig4} reports classification accuracy across $d \in \{8, 32, 100, 1000\}$ and
$\mathit{Cor} \in \{0, 0.2, 0.4, \allowbreak 0.5, 0.6, 0.8, \allowbreak 0.99\}$. The results show that the accuracy peaks at $Cor=0.5$ across all tested dimensions, which is consistent with our hypothesis. As $Cor$ increases from 0.0 to 0.5, correlated features better align with the entanglement pattern of the circuit. However, beyond $Cor = 0.6$, excessive correlation reduces feature diversity and compresses the effective input dimension, leading to performance degradation. These trends are stable across $\mu$, indicating that the observed optimum is not specific to a single data distribution. We therefore verify it on practical image datasets in the next section.

\section{Verification Using Experimental Data}
\label{sec:veri}

%-------------------------------------------------------------------------
\subsection{Experimental Data Setup}
\begin{figure}[t]
  \centering
  \includegraphics[width=0.8\linewidth]{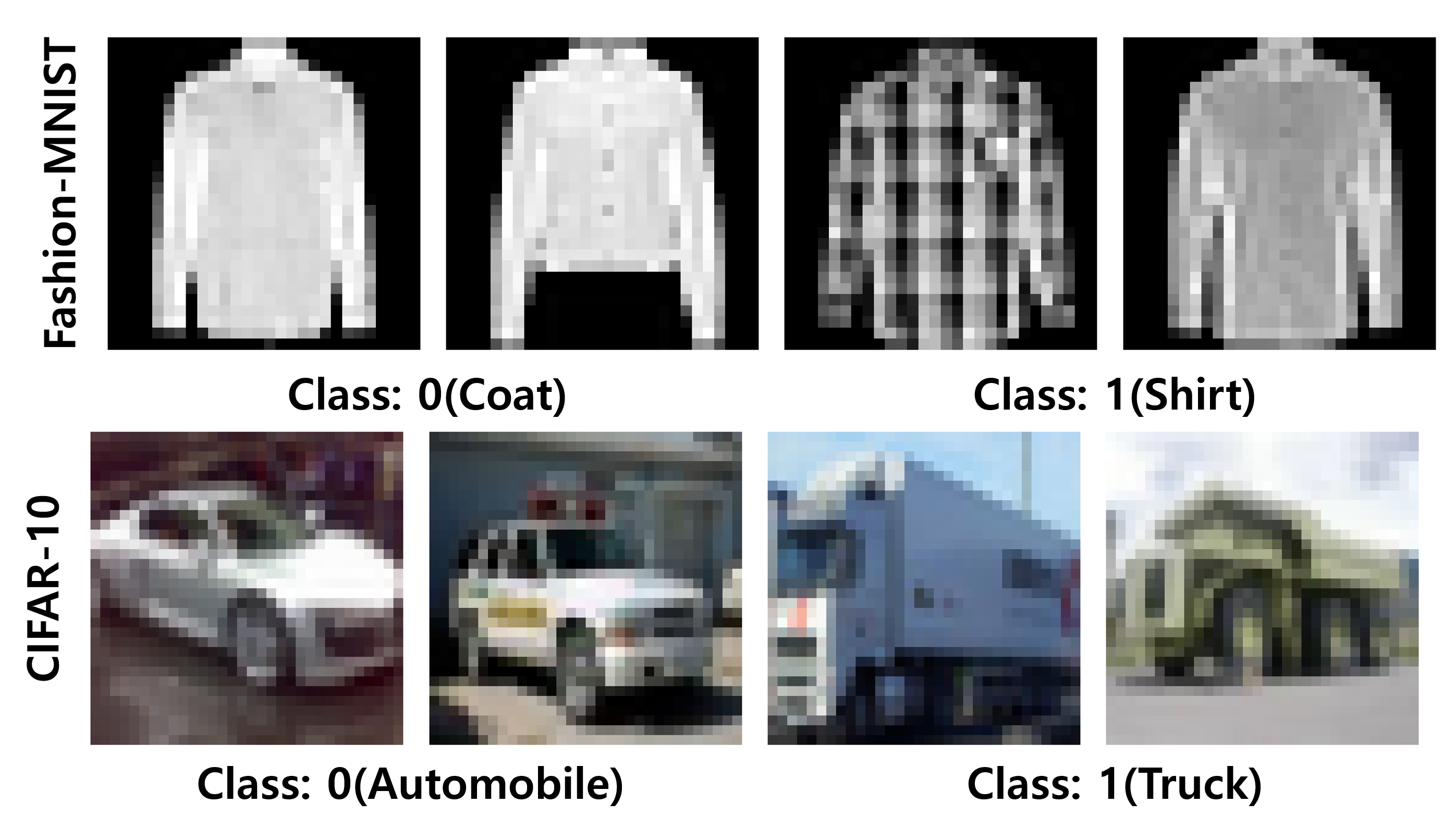}
  \caption{Examples for Fashion-MNIST and CIFAR-10.}
  \label{fig5}
\end{figure}

\begin{figure}[t]
  \centering
  % 1st row
  \subfloat[]{%
    \includegraphics[width=0.3\linewidth]{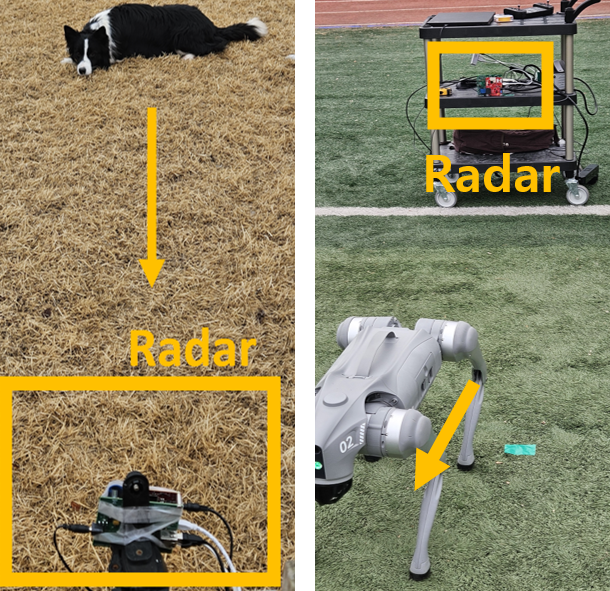}%
  }\hfill
  \subfloat[]{%
    \includegraphics[width=0.3\linewidth]{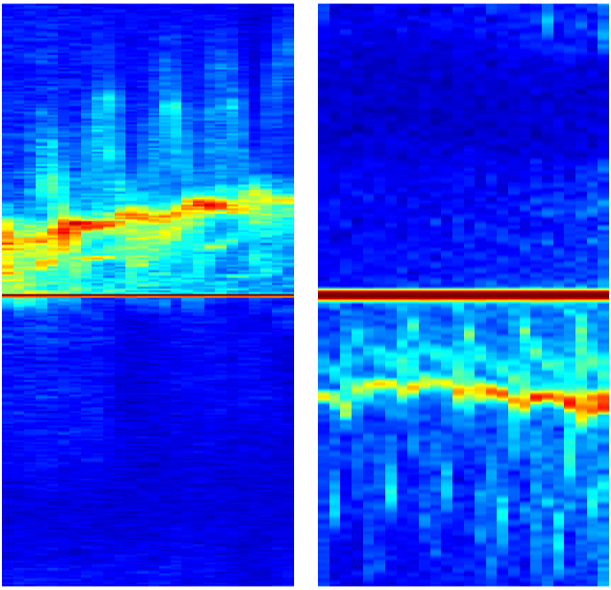}%
  }\hfill  
  \subfloat[]{%
    \includegraphics[width=0.3\linewidth]{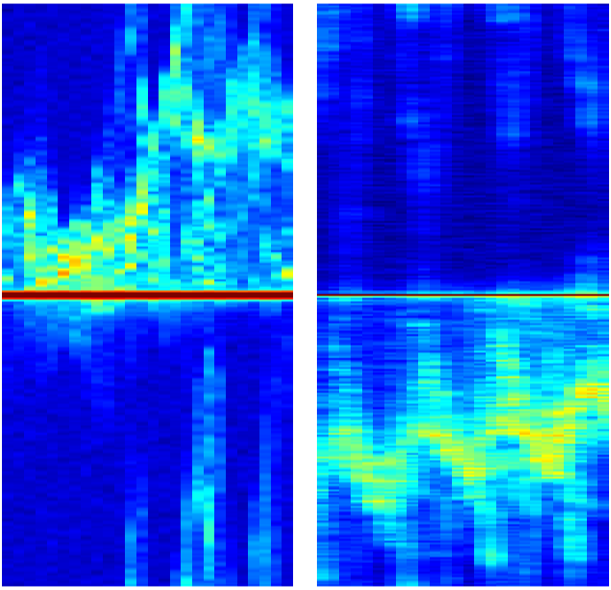}%
  }\\[0em]
  
  % 2nd row
  \subfloat[]{%
    \includegraphics[width=0.3\linewidth]{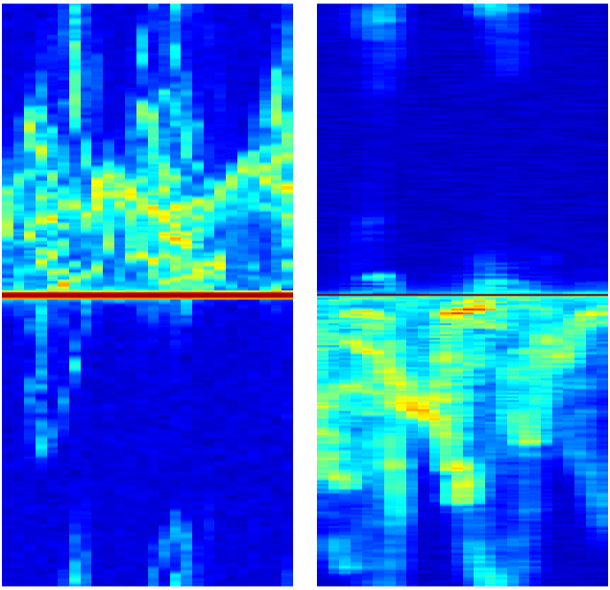}%
  }\hfill  
  \subfloat[]{%
    \includegraphics[width=0.3\linewidth]{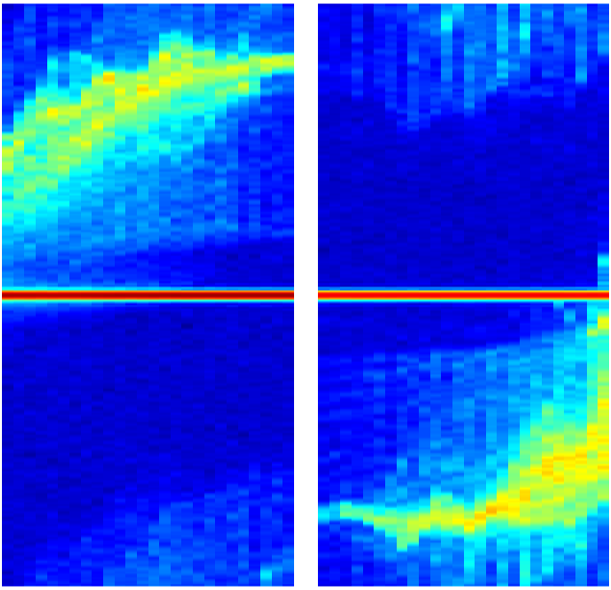}%
      }\hfill  
  \subfloat[]{%
    \includegraphics[width=0.3\linewidth]{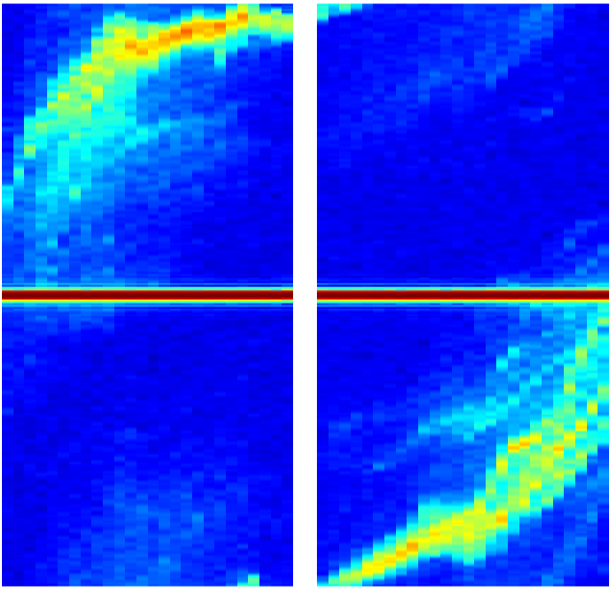}%
  }  
  \caption{Measurement setup and examples for micro-Doppler signature. (a)measurement setup, (b) a robotic dog, (c) a dog, (d) a walking person, (e) a person riding a bicycle, and (f) a person riding an electric scooter. The left signatures represent targets approaching the radar, while the right ones correspond to targets moving away from the radar. In (b)-(f), the vertical axis denotes Doppler frequency; the horizontal axis denotes time; color indicates spectrogram magnitude.}
  \label{fig6}
\end{figure}

We evaluate three binary classification tasks using the proposed correlation-regularized hybrid CNN--QNN: Fashion-MNIST (coat vs. shirt), CIFAR-10 (automobile vs. truck), and radar micro-Doppler signatures (robot vs. non-robot). For Fashion-MNIST and CIFAR-10, we selected the most confusing class pairs as depicted in \cref{fig5}. Micro-Doppler signatures are measured using frequency-modulated continuous-wave radar and spectrograms are generated by applying short-time Fourier transform. The Micro-Doppler signature refers to frequency shifts caused by micro-motions such as vibration, rotation, and limb motion \cite{chen2006micro}. Such micro-Doppler analysis has been widely used for target detection and classification \cite{zhao2019extraction, kwon2017human, kim2015human}. In this dataset, to classify robotic dogs and non-robots, a Unitree GO2 robotic dog and four non-robot targets—dogs, walking person, person riding a bicycle, and person riding an electric scooter—were measured as shown in \cref{fig6}. Experiments for micro-Doppler signatures involving human subjects were approved by an Institutional Review Board (IRB) at Sogang University (IRB Number: SGUIRB-A-2501-06). Input shapes and class sizes are: CIFAR-10 3×32×32, 6,000 samples per class; Fashion-MNIST 1×28×28, 7,000 per class; micro-Doppler 1×1500×26, 8,459 per class. All datasets are class-balanced. Each dataset is split 0.64:0.16:0.20 for training, validation, and testing.

As the CNN feature extractor, we investigate a shallow CNN with a single convolutional layer followed by rectified linear unit (ReLU) and adaptive average pooling layer, and ResNet-18. Both extract 8 features that are encoded onto 8 qubits. The CNN output is connected to the QNN classifier via a TorchConnector interface, and we jointly optimize all parameters with Adam to minimize $L=L_{\text{classification}}+L_{\text{correlation}}$. All quantum experiments are executed on a Qiskit simulator.
To study the effect of correlation control, we sweep the target correlation parameter $\mathit{Cor} \in \{0, 0.2, \allowbreak \dots, \allowbreak 0.99\}$, train the correlation regularized hybrid model, and evaluate classification performance. We also compare with a hybrid model not controlling correlations. We use 5-fold cross-validation and report the mean and standard deviation of accuracy across folds.

%-------------------------------------------------------------------------
\subsection{Classification Results}
The classification results are shown in \cref{tab:table1,tab:table2} Each table reports the results using a shallow CNN and  ResNet-18 as a  feature extractor, respectively. The left blocks of the table correspond to training with the loss function $L=L_{\text{classification}}+L_{\text{correlation}}$ as described in \cref{sec:method}, and its columns sweep the target correlation $Cor$. The rightmost columns labeled $L_{\text{classification}}$ are the baseline without the correlation regularization. For each dataset, we report the mean accuracy and standard deviation over five folds. Bold entries indicate the best mean accuracy within each dataset

Across backbones and datasets, it is shown that inducing a correlation $Cor\cong0.5$ improves classification accuracy and reduces standard deviation. The accuracy optimum concentrates near $Cor=0.5$, consistent with the results of \cref{sec:anal}, and the lowest standard deviation occurs within $Cor\in{[0.4, 0.6]}$ depending on the dataset. 

With the shallow CNN in \cref{tab:table1} on Fashion-MNIST, the model without correlation control achieves the highest accuracy which may appear to contradict our claim at first glance. However, this configuration is unstable: in 2 of 5 runs the model collapsed, producing an invariant output for all inputs. The performance shown in  \cref{tab:table1} was obtained only after re-training until a model trained successfully. On CIFAR-10, the unregularized baseline also collapsed in 1 of 5 runs. In contrast, adding the correlation loss prevented collapse and stabilized training, yielding repeatable results. Thus, the unregularized peak should be regarded as unreliable, while correlation control improves reliability even if its peak accuracy is slightly lower. 
With ResNet-18 in \cref{tab:table2}, classification performance improves over the shallow backbone on all datasets and no collapse is observed. For all datasets, the best performance occurs near $Cor\cong0.5$, which consistently yields the highest mean accuracy and the lowest standard deviation. 

To compare QNN performance against a classical head, we replace the QNN with a multilayer perceptron (MLP) classifier while keeping the same backbones. The MLP uses a ReLU followed by a fully connected output layer. With the shallow CNN, accuracy is 85.16\% $\pm$ 1.32 pp on Fashion-MNIST, 83.31\% $\pm$ 1.44 pp on CIFAR-10, and 90.40\% $\pm$ 2.17 pp on micro-Doppler. On the other hand, with ResNet-18, the corresponding figures are 94.24\% $\pm$ 0.62 pp, 93.50\% $\pm$ 0.31 pp, and 99.85\% $\pm$ 0.06 pp. These results indicate that, with a shallow feature extractor, the QNN head improves both classification accuracy and its variance relative to an MLP head. With a sufficiently strong extractor such as ResNet-18 the two heads achieve comparable performance.

\begin{table}[t]
\centering
\caption{Results comparison for a shallow CNN.}
\label{tab:table1}
\small % CVPR 표는 작게 표시 권장
\setlength{\tabcolsep}{4pt} % 열 간격 조정
\renewcommand{\arraystretch}{1.05} % 줄 간격 살짝 확장

% 폭 자동 맞춤 시작
\resizebox{\linewidth}{!}{
\begin{tabular}{@{}lcccccccc@{}}
\toprule
\multirow{3}{*}{\textbf{Loss function}} &
\multicolumn{7}{c}{$L = L_{\text{classification}} + L_{\text{correlation}}$} &
\multirow{3}{*}{$L_{\text{classification}}$} \\ 
\cmidrule(lr){2-8}

& \multicolumn{7}{c}{$\mathit{Cor}$} & \\ 
\cmidrule(lr){2-8}

& 0.0 & 0.2 & 0.4 & 0.5 & 0.6 & 0.8 & 1.0 & \\ 
\midrule

\multicolumn{9}{c}{\cellcolor[HTML]{EFEFEF}\textbf{Fashion-MNIST (shirt vs. coat)}} \\ \midrule
Acc. [\%] & 86.26 & 85.39 & 86.02 & 86.18 & 85.93 & 85.80 & 86.04 & \textbf{86.36} \\
Std. [pp] & 1.66  & 0.72  & 0.72  & 1.39  & 0.70  & 1.63  & 0.75  & \makecell{\textbf{0.67} \\ \scriptsize (unstable)} \\ \midrule

\multicolumn{9}{c}{\cellcolor[HTML]{EFEFEF}\textbf{CIFAR-10 (truck vs. automobile)}} \\ \midrule
Acc. [\%] & 85.10 & 85.19 & \textbf{86.00} & 85.60 & 85.15 & 85.80 & 85.10 & 85.48 \\
Std. [pp] & 0.57  & 0.71  & \textbf{0.57}  & 0.57  & 0.51  & 1.63  & 0.76  & \makecell{0.60 \\ \scriptsize (unstable)} \\ \midrule

\multicolumn{9}{c}{\cellcolor[HTML]{EFEFEF}\textbf{Radar micro-Doppler (robot vs. non-robot)}} \\ \midrule
Acc. [\%] & 93.75 & 94.22 & 94.56 & \textbf{95.33} & \textbf{95.09} & 94.30 & 94.61 & 92.21 \\
Std. [pp] & 1.84  & 1.85  & 0.90  & \textbf{0.96}  & \textbf{0.58}  & 0.40  & 0.61  & 3.59 \\ 
\bottomrule
\end{tabular}
} % 폭 자동 맞춤 끝
\end{table}

\begin{table}[t]
\centering
\caption{Results comparison for ResNet-18.}
\label{tab:table2}
\small 
\setlength{\tabcolsep}{4pt} 
\renewcommand{\arraystretch}{1.05} 

\resizebox{\linewidth}{!}{
\begin{tabular}{@{}lcccccccc@{}}
\toprule
\multirow{3}{*}{\textbf{Loss function}} &
\multicolumn{7}{c}{$L = L_{\text{classification}} + L_{\text{correlation}}$} &
\multirow{3}{*}{$L_{\text{classification}}$} \\ 
\cmidrule(lr){2-8}

& \multicolumn{7}{c}{$\mathit{Cor}$} & \\ 
\cmidrule(lr){2-8}

& 0.0 & 0.2 & 0.4 & 0.5 & 0.6 & 0.8 & 1.0 & \\ 
\midrule

\multicolumn{9}{c}{\cellcolor[HTML]{EFEFEF}\textbf{Fashion-MNIST (shirt vs. coat)}} \\ \midrule
Acc. [\%] & 91.97 & 92.11 & 92.16 & \textbf{92.96} & 91.51 & 92.56 & 92.11 & 91.24 \\
Std. [pp] & 1.25  & 2.51  & 2.19  & \textbf{0.99}  & 2.47  & 2.61  & 2.80  & 2.40 \\ \midrule

\multicolumn{9}{c}{\cellcolor[HTML]{EFEFEF}\textbf{CIFAR-10 (truck vs. automobile)}} \\ \midrule
Acc. [\%] & 91.15 & 90.65 & 91.67 & 91.33 & \textbf{92.13} & 91.34 & 92.04 & 91.53 \\
Std. [pp] & 0.87  & 0.42  & 0.67  & 0.54  & \textbf{1.05}  & 1.07  & 0.49  & 0.91 \\ \midrule

\multicolumn{9}{c}{\cellcolor[HTML]{EFEFEF}\textbf{Radar micro-Doppler (robot vs. non-robot)}} \\ \midrule
Acc. [\%] & 99.41 & 99.63 & 99.52 & \textbf{99.65} & \textbf{99.63} & 99.61 & 99.33 & 99.57 \\
Std. [pp] & 0.26  & 0.17  & 0.21  & \textbf{0.11}  & \textbf{0.06}  & 0.27  & 0.18  & 0.15 \\
\bottomrule
\end{tabular}
} 
\end{table}
\section{Conclusion}
\label{sec:con}

We confirmed that controlling classical feature correlations to better utilize the QNN's entangling capacity improves classification performance under shallow, qubit-limited VQC settings. We proposed a simple correlation-regularization term that steers the off-diagonal entries of the feature correlation matrix toward a target $\mathit{Cor}$, and incorporate it to a hybrid CNN--QNN pipeline. We presented a mathematical derivation that found the optimum correlation value to maximize classification accuracy. Three binary tasks (Fashion-MNIST, CIFAR-10, radar micro-Doppler) were evaluated using a Qiskit simulator with 5-fold cross-validation. The experimental results were consistent with the simulation, indicating that a moderate level of correlation is beneficial.  Accuracy peaks near $Cor\cong0.5$ across backbones and datasets, and variance is minimized in the range of $\mathit{Cor}\in [0.4, 0.6]$. Notably, the conventional practice of inputting orthogonal--–decorrelated--–features to models was not optimal in quantum models; an intermediate correlation level exceeded the baseline's accuracy while reducing variance. This result suggests that QNN can outperform classical classifiers as more qubits become available, as this study was limited to eight qubits.

Operationally, the proposed method requires no circuit changes and only introduces one scalar hyperparameter $Cor$ for feature extraction. A practical design rule follows: before quantum encoding, induce intermediate feature correlation, defaulting to $\mathit{Cor} \cong 0.5$, and fine-tune within 0.4--0.6 to balance accuracy and stability. 

Limitations include binary classification, simulator-based experiments due to the resource constraints, and the use of fixed uniform correlation targets. Future directions involve extending the framework to multiclass tasks, evaluating on real hardware with noise, and exploring various forms of VQCs and target correlation matrices.

\section*{Acknowledgments}
This work was supported by Institute of Information \& Communications Technology Planning \& Evaluation (IITP) grant funded by the Korea government (MSIT) (No. RS-2024-00393808, Efficient design of RF components and systems based on artificial intelligence, 100\%) and National Research Foundation (NRF), Republic of Korea [BK21 FOUR].

{
    \small
    \bibliographystyle{ieeenat_fullname}

    \bibliography{main}

@article{preskill2018quantum,
  title={Quantum computing in the NISQ era and beyond},
  author={Preskill, John},
  journal={Quantum},
  volume={2},
  pages={79},
  year={2018},
  publisher={Verein zur F{\"o}rderung des Open Access Publizierens in den Quantenwissenschaften}
}

@article{cheng2021simulating,
  title={Simulating noisy quantum circuits with matrix product density operators},
  author={Cheng, Song and Cao, Chenfeng and Zhang, Chao and Liu, Yongxiang and Hou, Shi-Yao and Xu, Pengxiang and Zeng, Bei},
  journal={Physical review research},
  volume={3},
  number={2},
  pages={023005},
  year={2021},
  publisher={APS}
}

@article{mcclean2018barren,
  title={Barren plateaus in quantum neural network training landscapes},
  author={McClean, Jarrod R and Boixo, Sergio and Smelyanskiy, Vadim N and Babbush, Ryan and Neven, Hartmut},
  journal={Nature communications},
  volume={9},
  number={1},
  pages={4812},
  year={2018},
  publisher={Nature Publishing Group UK London}
}

@article{sim2019expressibility,
  title={Expressibility and entangling capability of parameterized quantum circuits for hybrid quantum-classical algorithms},
  author={Sim, Sukin and Johnson, Peter D and Aspuru-Guzik, Al{\'a}n},
  journal={Advanced Quantum Technologies},
  volume={2},
  number={12},
  pages={1900070},
  year={2019},
  publisher={Wiley Online Library}
}

@article{jerbi2023quantum,
  title={Quantum machine learning beyond kernel methods},
  author={Jerbi, Sofiene and Fiderer, Lukas J and Poulsen Nautrup, Hendrik and K{\"u}bler, Jonas M and Briegel, Hans J and Dunjko, Vedran},
  journal={Nature Communications},
  volume={14},
  number={1},
  pages={517},
  year={2023},
  publisher={Nature Publishing Group UK London}
}

@article{hur2022quantum,
  title={Quantum convolutional neural network for classical data classification},
  author={Hur, Tak and Kim, Leeseok and Park, Daniel K},
  journal={Quantum Machine Intelligence},
  volume={4},
  number={1},
  pages={3},
  year={2022},
  publisher={Springer}
}

@article{gong2024quantum,
  title={Quantum convolutional neural network based on variational quantum circuits},
  author={Gong, Li-Hua and Pei, Jun-Jie and Zhang, Tian-Feng and Zhou, Nan-Run},
  journal={Optics Communications},
  volume={550},
  pages={129993},
  year={2024},
  publisher={Elsevier}
}

@article{wang2024transition,
  title={Transition role of entangled data in quantum machine learning},
  author={Wang, Xinbiao and Du, Yuxuan and Tu, Zhuozhuo and Luo, Yong and Yuan, Xiao and Tao, Dacheng},
  journal={Nature Communications},
  volume={15},
  number={1},
  pages={3716},
  year={2024},
  publisher={Nature Publishing Group UK London}
}

@inproceedings{falez2020improving,
  title={Improving stdp-based visual feature learning with whitening},
  author={Falez, Pierre and Tirilly, Pierre and Bilasco, Ioan Marius},
  booktitle={2020 International Joint Conference on Neural Networks (IJCNN)},
  pages={1--8},
  year={2020},
  organization={IEEE}
}

@inproceedings{pan2019switchable,
  title={Switchable whitening for deep representation learning},
  author={Pan, Xingang and Zhan, Xiaohang and Shi, Jianping and Tang, Xiaoou and Luo, Ping},
  booktitle={Proceedings of the IEEE/CVF international conference on computer vision},
  pages={1863--1871},
  year={2019}
}

@inproceedings{zbontar2021barlow,
  title={Barlow twins: Self-supervised learning via redundancy reduction},
  author={Zbontar, Jure and Jing, Li and Misra, Ishan and LeCun, Yann and Deny, St{\'e}phane},
  booktitle={International conference on machine learning},
  pages={12310--12320},
  year={2021},
  organization={PMLR}
}

@article{cogswell2015reducing,
  title={Reducing overfitting in deep networks by decorrelating representations},
  author={Cogswell, Michael and Ahmed, Faruk and Girshick, Ross and Zitnick, Larry and Batra, Dhruv},
  journal={arXiv preprint arXiv:1511.06068},
  year={2015}
}

@article{rodriguez2016regularizing,
  title={Regularizing cnns with locally constrained decorrelations},
  author={Rodr{\'\i}guez, Pau and Gonzalez, Jordi and Cucurull, Guillem and Gonfaus, Josep M and Roca, Xavier},
  journal={arXiv preprint arXiv:1611.01967},
  year={2016}
}

@article{hafeez2024h,
  title={H-QNN: A hybrid quantum--classical neural network for improved binary image classification},
  author={Hafeez, Muhammad Asfand and Munir, Arslan and Ullah, Hayat},
  journal={AI},
  volume={5},
  number={3},
  pages={1462--1481},
  year={2024},
  publisher={MDPI}
}

@article{chalumuri2022quantum,
  title={Quantum--classical image processing for scene classification},
  author={Chalumuri, Avinash and Kune, Raghavendra and Kannan, S and Manoj, BS},
  journal={IEEE Sensors Letters},
  volume={6},
  number={6},
  pages={1--4},
  year={2022},
  publisher={IEEE}
}

@article{fan2023hybrid,
  title={Hybrid quantum-classical convolutional neural network model for image classification},
  author={Fan, Fan and Shi, Yilei and Guggemos, Tobias and Zhu, Xiao Xiang},
  journal={IEEE transactions on neural networks and learning systems},
  year={2023},
  publisher={IEEE}
}

@inproceedings{ray2024hybrid,
  title={Hybrid quantum-classical graph neural networks for tumor classification in digital pathology},
  author={Ray, Anupama and Madan, Dhiraj and Patil, Srushti and Pati, Pushpak and Rapsomaniki, Marianna and Kohlakala, Aviwe and Dlamini, Thembelihle Rose and Muller, Stephanie Julia and Rhrissorrakrai, Kahn and Utro, Filippo and others},
  booktitle={2024 IEEE international conference on Quantum Computing and Engineering (QCE)},
  volume={1},
  pages={1611--1616},
  year={2024},
  organization={IEEE}
}

@inproceedings{afane2025atp,
  title={ATP: Adaptive Threshold Pruning for Efficient Data Encoding in Quantum Neural Networks},
  author={Afane, Mohamed and Ebbrecht, Gabrielle and Wang, Ying and Chen, Juntao and Farooq, Junaid},
  booktitle={Proceedings of the Computer Vision and Pattern Recognition Conference},
  pages={20427--20436},
  year={2025}
}

@article{choudhary2025hqnn,
  title={HQNN-FSP: A Hybrid Classical-Quantum Neural Network for Regression-Based Financial Stock Market Prediction},
  author={Choudhary, Prashant Kumar and Innan, Nouhaila and Shafique, Muhammad and Singh, Rajeev},
  journal={arXiv preprint arXiv:2503.15403},
  year={2025}
}

@inproceedings{ghosh2024hybrid,
  title={A Hybrid Quantum-Classical CNN Architecture for Semantic Segmentation of Radar Sounder Data},
  author={Ghosh, Raktim and Delilbasic, Amer and Cavallaro, Gabriele and Bovolo, Francesca},
  booktitle={2024 IEEE Mediterranean and Middle-East Geoscience and Remote Sensing Symposium (M2GARSS)},
  pages={366--370},
  year={2024},
  organization={IEEE}
}

@article{liu2025radar,
  title={Radar HRRP Target Recognition Based on Hybrid Quantum Neural Networks},
  author={Liu, Xin and Zhou, Daiying and Huang, Qiuyan},
  journal={IEEE Transactions on Aerospace and Electronic Systems},
  year={2025},
  publisher={IEEE}
}

@article{le2011flexible,
  title={A flexible representation of quantum images for polynomial preparation, image compression, and processing operations},
  author={Le, Phuc Q and Dong, Fangyan and Hirota, Kaoru},
  journal={Quantum Information Processing},
  volume={10},
  number={1},
  pages={63--84},
  year={2011},
  publisher={Springer}
}

@article{schuld2019evaluating,
  title={Evaluating analytic gradients on quantum hardware},
  author={Schuld, Maria and Bergholm, Ville and Gogolin, Christian and Izaac, Josh and Killoran, Nathan},
  journal={Physical Review A},
  volume={99},
  number={3},
  pages={032331},
  year={2019},
  publisher={APS}
}

@article{havlivcek2019supervised,
  title={Supervised learning with quantum-enhanced feature spaces},
  author={Havl{\'\i}{\v{c}}ek, Vojt{\v{e}}ch and C{\'o}rcoles, Antonio D and Temme, Kristan and Harrow, Aram W and Kandala, Abhinav and Chow, Jerry M and Gambetta, Jay M},
  journal={Nature},
  volume={567},
  number={7747},
  pages={209--212},
  year={2019},
  publisher={Nature Publishing Group UK London}
}

@article{chen2006micro,
  title={Micro-Doppler effect in radar: phenomenon, model, and simulation study},
  author={Chen, Victor C and Li, Fayin and Ho, S-S and Wechsler, Harry},
  journal={IEEE Transactions on Aerospace and electronic systems},
  volume={42},
  number={1},
  pages={2--21},
  year={2006},
  publisher={IEEE}
}

@article{zhao2019extraction,
  title={The extraction of micro-Doppler signal with EMD algorithm for radar-based small UAVs’ detection},
  author={Zhao, Yichao and Su, Yi},
  journal={IEEE Transactions on Instrumentation and Measurement},
  volume={69},
  number={3},
  pages={929--940},
  year={2019},
  publisher={IEEE}
}

@inproceedings{kwon2017human,
  title={Human detection by neural networks using a low-cost short-range Doppler radar sensor},
  author={Kwon, Jihoon and Kwak, Nojun},
  booktitle={2017 IEEE Radar Conference (RadarConf)},
  pages={0755--0760},
  year={2017},
  organization={IEEE}
}

@article{kim2015human,
  title={Human detection and activity classification based on micro-Doppler signatures using deep convolutional neural networks},
  author={Kim, Youngwook and Moon, Taesup},
  journal={IEEE geoscience and remote sensing letters},
  volume={13},
  number={1},
  pages={8--12},
  year={2015},
  publisher={IEEE}
}

@article{garriga2018deep,
  title={Deep convolutional networks as shallow gaussian processes},
  author={Garriga-Alonso, Adri{\`a} and Rasmussen, Carl Edward and Aitchison, Laurence},
  journal={arXiv preprint arXiv:1808.05587},
  year={2018}
}

@book{nielsen2010quantum,
  title={Quantum computation and quantum information},
  author={Nielsen, Michael A and Chuang, Isaac L},
  year={2010},
  publisher={Cambridge university press}
}

}


\end{document}